\documentclass[conference]{IEEEtran}
\usepackage{xcolor,soul,framed} 

\colorlet{shadecolor}{yellow}
\usepackage[pdftex]{graphicx}
\graphicspath{{../pdf/}{../jpeg/}}
\DeclareGraphicsExtensions{.pdf,.jpeg,.png}

\usepackage[cmex10]{amsmath}
\usepackage{cite}
\usepackage{flushend}
\usepackage{rotating}
\usepackage{hyperref} 
\usepackage{tablefootnote}
\usepackage{array}
\usepackage{mdwmath}
\usepackage{amsmath}
\usepackage{amsfonts}
\usepackage{amssymb}
\usepackage{mdwtab}
\usepackage{eqparbox}
\usepackage{url}
\usepackage{xcolor}
\usepackage{comment}
\usepackage{footnote}
\usepackage{multirow}
\usepackage{graphicx}
\usepackage{caption}
\usepackage{algorithm}
\usepackage{algorithmic}
\allowdisplaybreaks

\usepackage{float}
\usepackage{placeins}

\usepackage{url} 
\usepackage{array, makecell}

\usepackage{tikz}

\usetikzlibrary{fit,positioning}
\usetikzlibrary{bayesnet}
\usetikzlibrary{arrows}

\begin{document}
\bstctlcite{IEEEexample:BSTcontrol}
    \title{Semantic Knowledge Distillation for Onboard Satellite Earth Observation Image Classification}

    \vspace{-2mm}
    
    \author{\IEEEauthorblockN{Thanh-Dung Le, Vu Nguyen Ha, Ti Ti Nguyen, Geoffrey Eappen, Prabhu Thiruvasagam, \\
    Hong-fu Chou, Duc-Dung Tran, Luis M. Garces-Socarras, Jorge L. Gonzalez-Rios,\\
    Juan Carlos Merlano-Duncan, Symeon Chatzinotas 
    }\\
    \vspace{-2mm}
    \IEEEauthorblockA{\textit{Interdisciplinary Centre for Security, Reliability and Trust (SnT), University of Luxembourg, Luxembourg} 
	}
     \vspace{-8mm}
    }
\markboth{IEEE, VOL., NO., 2024.
}{Thanh-Dung Le \MakeLowercase{\textit{et al.}}: }

 \maketitle

\begin{abstract}
This study introduces a dynamic weighting knowledge distillation (KD) framework for efficient Earth observation (EO) image classification (IC) in resource-constrained environments. By leveraging EfficientViT and MobileViT as teacher models, this approach enables lightweight student models, specifically ResNet8 and ResNet16, to achieve over 90\% accuracy, precision, and recall, meeting the confidence thresholds required for reliable classification. Unlike traditional KD with fixed weights, our dynamic weighting mechanism adjusts based on each teacher’s confidence, allowing the student model to prioritize more reliable knowledge sources. ResNet8, in particular, achieves substantial efficiency gains, with 97.5\% fewer parameters, 96.7\% fewer FLOPs, 86.2\% lower power consumption, and 63.5\% faster inference time compared to MobileViT. This significant reduction in complexity and resource demand makes ResNet8 an ideal choice for EO tasks, balancing high performance with practical deployment requirements. This confidence-driven, adaptable KD strategy demonstrates the potential of dynamic knowledge distillation to deliver high-performing, resource-efficient models for satellite-based EO applications. Reproducible codes are available from our shared Github repository \footnote{
\scriptsize
\url{https://github.com/ltdung/SnT-SENTRY}}.

\end{abstract}


\IEEEpeerreviewmaketitle


\section{Introduction}

\IEEEPARstart{T}{he} rapid increase in satellite deployments for EO and remote sensing (RS) missions reflects a growing demand for applications like environmental monitoring, disaster response, precision agriculture, and scientific research \cite{sadek2021new}. These applications rely on high-frequency, high-resolution data for timely and accurate decision-making. However, a significant bottleneck in Low Earth Orbit (LEO) satellite operations is reliance on ground stations for data transmission, which limits the availability of communication windows and results in frequent connectivity loss \cite{al2022survey}. This delay can impede critical responses in situations requiring immediate data access.

The advent of Satellite Internet Providers, such as Starlink and OneWeb, offers the potential for continuous (24/7) connectivity to LEO satellites, facilitating on-demand data access \cite{chougrani2024connecting}. Yet, seamless connectivity alone does not fully meet modern EO and RS requirements, which increasingly demand real-time, onboard decision-making. For optimal operations, onboard neural networks (NNs) must prioritize computational efficiency to autonomously analyze data, identify critical information, and make immediate adjustments, such as refocusing on a target area during subsequent satellite passes \cite{fontanesi2023artificial}.

Historically, onboard NNs have been designed for efficiency, often relying on convolutional neural network (CNN) models to balance performance and resource constraints. For example, the $\Phi$-Sat-1 mission used a CNN-based NN for onboard image segmentation using the Intel Movidius Myriad 2 vision processing unit (VPU), representing the first deployment of deep learning on a satellite \cite{giuffrida2021varphi}. Similarly, $\Phi$-Sat-2 adopted a convolutional autoencoder for image compression to reduce transmission requirements, demonstrating the feasibility of lightweight models on hardware-constrained environments on three different hardware, including graphic processing unit (GPU) NVIDIA GeForce GTX 1650, VPU Myriad 2, and central processing unit (CPU) Intel Core i7-6700\cite{guerrisi2023artificial}. 

Despite their efficiency, CNNs can be limited in performance, especially compared to the recent success of Vision Transformer (ViT) architectures. ViTs have gained popularity in computer vision due to their ability to capture global context via self-attention mechanisms, often surpassing traditional CNNs in performance. However, ViTs require significantly more computational power and memory as image resolution increases, which poses challenges for deployment on power-constrained satellite platforms \cite{chou2024semantic, le2024board}.

To overcome these limitations, KD offers a viable approach for onboard processing. KD is a method where a smaller, simpler model (the student) learns from a larger, complex model (the teacher, such as a ViT). By transferring the teacher's semantic knowledge, KD allows the student to generalize more effectively with lower computational demands \cite{papa2024survey}. KD was initially introduced to reduce the computational burden of deep learning models \cite{hinton2015distilling}, and recent studies indicate that KD can help students learn complex representations with strong performance even in simplified forms \cite{stanton2021does}.

In this study, we leverage KD to train deployable models for onboard EO tasks, explicitly focusing on IC. By distilling semantic knowledge from ViTs into efficient student models, we aim to boost onboard processing capabilities while maintaining computational efficiency suitable for satellite EO missions. Traditional KD approaches often struggle with training instability, mainly when exact prediction matches are enforced through Kullback-Leibler (KL) divergence from a single teacher, which can impair performance \cite{huang2022knowledge}. To address this, we propose a dynamic weighting mechanism for dual-teacher KD (DualKD), where the weight assigned to each teacher adapts based on their confidence level, enabling the student model to prioritize the most reliable knowledge sources. This approach considers that one instance may have varying semantic similarities to different teachers, thus improving the student’s ability to generalize data representation.

\begin{figure*}[!ht]
    \centering
    \includegraphics[scale=0.325]{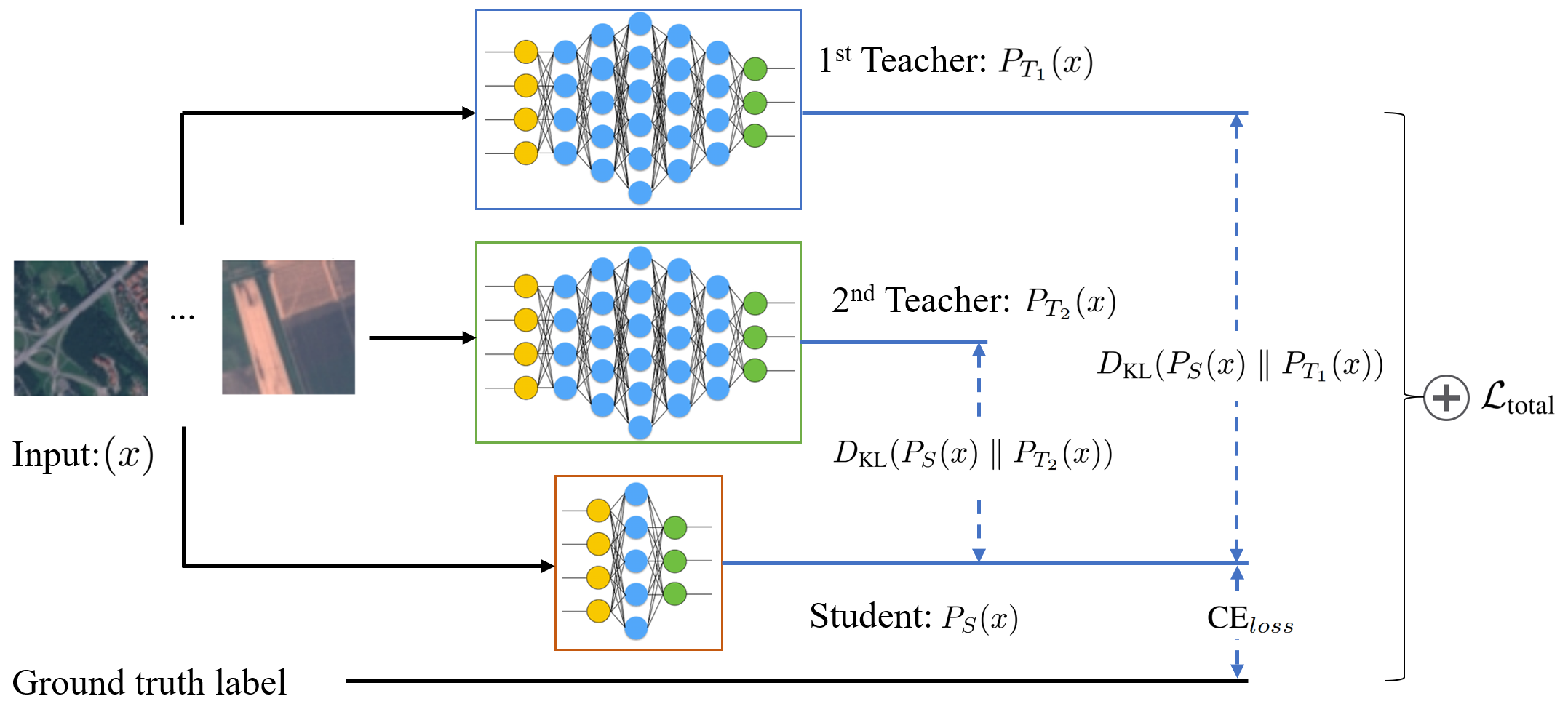}
    \captionsetup{font=small}
    \caption{The schematic workflow of DualKD.}
    \label{fig:workflow}
    \vspace{-4mm}
\end{figure*} 

\section{Materials and Methods}
\subsection{Dataset}

This study utilizes the EuroSAT dataset \cite{helber2019eurosat}, a well-established land use and land cover classification benchmark specifically curated from Sentinel-2 satellite imagery. EuroSAT includes 27,000 labeled, geo-referenced images, each with a resolution of 64x64 pixels across 13 spectral bands, and is organized into 10 distinct classes. These classes cover various land types, such as industrial and residential buildings, annual and permanent crops, rivers, lakes, herbaceous vegetation, highways, pastures, and forests, representing diverse European landscapes. Each class contains between 2000 and 3000 images, providing a balanced model training and evaluation dataset. EuroSAT’s compact image size and broad class diversity make it suitable for developing and assessing deep learning models intended for onboard processing in EO missions. This dataset is valuable for applications that demand real-time decision-making capabilities, such as environmental monitoring, disaster response, and precision agriculture.

\subsection{Dual Teachers Knowledge Distillation}

Traditional KD methods often face training instability, especially when forced to closely match a single teacher model’s predictions using KL divergence, which can compromise performance \cite{huang2022knowledge}. We propose a DualKD framework with a dynamic weighting mechanism to overcome this challenge and enhance the student model’s adaptability and effectiveness. Unlike traditional KD approaches that employ fixed weights, this method adjusts each teacher's influence based on confidence levels, allowing the student model to prioritize knowledge from the more reliable teacher. This dynamic weighting approach improves flexibility in handling multiple teachers with varying degrees of reliability, ultimately optimizing the semantic information from the KD process.

As shown in Fig. \ref{fig:workflow}, given an input $x$, the semantic distillation process starts by computing softened probability distributions for the teacher models and the student model. This is achieved by scaling their logits with a temperature parameter $\tau$. For teacher model $T_1$, the softened probability distribution is:
\begin{equation}
    P_{T_1}(x) = \text{softmax}\left(\frac{T_1(x)}{\tau}\right),
\end{equation}
and similarly for teacher model $T_2$:
\begin{equation}
    P_{T_2}(x) = \text{softmax}\left(\frac{T_2(x)}{\tau}\right),
\end{equation}
with the student model $S$:
\begin{equation}
    P_S(x) = \text{softmax}\left(\frac{S(x)}{\tau}\right).
\end{equation}

Confidence for each teacher is computed as the average of the maximum probabilities in their respective softened distributions:
\begin{equation}
    C_{T_1} = \mathbb{E}\left[\max(P_{T_1}(x))\right], \quad C_{T_2} = \mathbb{E}\left[\max(P_{T_2}(x))\right].
\end{equation}
Based on these confidence scores, we dynamically adjust the weights $\alpha$ and $\beta$ assigned to each teacher in the distillation loss $\text{KD}_{\text{loss}}$. If both confidence scores are significantly below a predefined threshold $\delta$, both teachers are ignored ($\alpha = \beta = 0$). If either confidence score is close to the threshold, we prioritize the more reliable teacher by reducing the weight of the less reliable one, with minimum weights set by $w_{min}$. When both teachers are above the threshold, equal weights ($\alpha = \beta = 0.5$) are used.

The distillation loss $\text{KD}_{loss}$, a weighted KL divergence between the student’s and each teacher’s softened probabilities, is then computed, with the weights $\alpha$ and $\beta$ reflecting each teacher’s confidence.

\begin{equation}
    \text{KD}_{\text{loss}} = \alpha \cdot D_{\text{KL}}(P_S(x) \parallel P_{T_1}(x)) + \beta \cdot D_{\text{KL}}(P_S(x) \parallel P_{T_2}(x)),
\end{equation}

Where the KL divergence $D_{\text{KL}}$ for each teacher-student pair is scaled by the temperature squared, $\tau^2$, to stabilize training:

\begin{equation}
    D_{\text{KL}}(P_S(x) \parallel P_{T_i}(x)) = \frac{1}{\tau^2} \sum_{j} P_{T_i}(x)_j \log\left(\frac{P_{T_i}(x)_j}{P_S(x)_j}\right)
\end{equation}

The total distillation loss is calculated as a combination of the classification loss, $\text{CE}_{\text{loss}}$, and the distillation loss $\text{KD}_{\text{loss}}$. A classification loss $\text{CE}_{loss}$ between the student’s predictions and the true labels is calculated to ground the student’s learning in teacher guidance and actual labels, where:

\begin{equation}
    \text{CE}_{\text{loss}} = - \sum_{i} y_i \log \left( P_S(x)_i \right),
\end{equation}

Then, the final combined loss, $\mathcal{L}_{\text{total}}$, integrates these components: a weighted combination of the $\text{CE}_{loss}$ and the $\text{KD}_{loss}$. This framework allows the student to leverage insights from both teachers selectively, focusing on the most reliable sources for improved generalization and adaptability across instances during training.

\begin{equation}
    \mathcal{L}_{\text{total}} = \left(1 - \frac{\alpha + \beta}{2}\right) \cdot \text{CE}_{\text{loss}} + \frac{\alpha + \beta}{2} \cdot \text{KD}_{\text{loss}}
\end{equation}

\begin{algorithm}
\caption{DualKD with Dynamic Weighting}
\label{alg:dual_teacher_distillation}
\footnotesize
\begin{algorithmic}[1]
\REQUIRE Input data $x$, true labels $y$, $\text{model\_student}$, $\text{model\_teacher\_1}$, $\text{model\_teacher\_2}$, temperature $\tau$, confidence threshold ($\delta$), minimum weight ($w_{min}$)
\ENSURE Combined loss $\mathcal{L}_{total}$ for backpropagation

\STATE \textbf{Forward pass:}
\STATE \quad $T_1(x) \leftarrow \text{model\_teacher\_1}(x)$
\STATE \quad $T_2(x) \leftarrow \text{model\_teacher\_2}(x)$
\STATE \quad $S(x) \leftarrow \text{model\_student}(x)$

\STATE \textbf{Calculate softened probabilities with temperature} $\tau$:
\STATE \quad $P_{T_1}(x) \leftarrow \text{softmax}(T_1(x) / \tau)$
\STATE \quad $P_{T_2}(x) \leftarrow \text{softmax}(T_2(x) / \tau)$
\STATE \quad $P_S(x) \leftarrow \text{softmax}(S(x) / \tau)$

\STATE \textbf{Calculate confidence scores for both teachers:}
\STATE \quad $C_{T_1} \leftarrow \mathbb{E} \left[\max(P_{T_1}(x))\right]$
\STATE \quad $C_{T_2} \leftarrow \mathbb{E} \left[\max(P_{T_2}(x))\right]$

\STATE \textbf{Dynamically set weights} $\alpha$ \textbf{and} $\beta$ \textbf{based on confidence scores:}
\IF{$C_{T_1} < 0.4$ \textbf{and} $C_{T_2} < 0.4$}
    \STATE $\alpha, \beta \leftarrow 0.0, 0.0$  \hfill // Ignore both teachers
\ELSIF{$C_{T_1} < \delta$ \textbf{and} $C_{T_2} < \delta$}
    \STATE $\alpha \leftarrow \max(0.5 - (\delta - C_{T_1}), w_{min})$
    \STATE $\beta \leftarrow \max(0.5 - (\delta - C_{T_2}), w_{min})$
\ELSIF{$C_{T_1} < \delta$}
    \STATE $\alpha, \beta \leftarrow 0.3, 0.7$  \hfill // Reduce $\alpha$, prioritize teacher 2
\ELSIF{$C_{T_2} < \delta$}
    \STATE $\alpha, \beta \leftarrow 0.7, 0.3$  \hfill // Reduce $\beta$, prioritize teacher 1
\ELSE
    \STATE $\alpha, \beta \leftarrow 0.5, 0.5$  \hfill // Equal weighting for both confident teachers
\ENDIF

\STATE \textbf{Compute Distillation Loss (KL Divergence with weighted sum):}
\STATE \quad $\text{loss}_1 \leftarrow D_{KL}(P_S(x) \parallel P_{T_1}(x))$
\STATE \quad $\text{loss}_2 \leftarrow D_{KL}(P_S(x) \parallel P_{T_2}(x))$
\STATE \quad $\text{KD}_{loss} \leftarrow (\alpha \cdot \text{loss}_1 + \beta \cdot \text{loss}_2) \cdot \tau^2$

\STATE \textbf{Compute Classification Loss (Cross-Entropy):}
\STATE \quad $\text{CE}_{loss} \leftarrow - \sum_{i} y_i \log \left( P_S(x)_i \right)$

\STATE \textbf{Combine losses to calculate Total Loss:}
\STATE \quad $\mathcal{L}_{total} \leftarrow \left(1 - \frac{\alpha + \beta}{2}\right) \cdot \text{CE}_{loss} + \frac{\alpha + \beta}{2} \cdot \text{KD}_{loss}$

\STATE \textbf{Backpropagate using} $\mathcal{L}_{total}$

\end{algorithmic}
\normalsize
\end{algorithm}

The pseudo Algorithm \ref{alg:dual_teacher_distillation} presents the core steps for implementing the proposed DualKD framework with a dynamic weighting mechanism that prioritizes the semantic knowledge from two teacher models. This adaptive approach is designed to maximize the student model’s learning efficiency by allowing it to concentrate on information from the most reliable teacher in each instance. Specifically, each teacher model's confidence score is computed by averaging the maximum probabilities of their softened outputs across a batch, reflecting each teacher’s reliability. These scores are then used to dynamically adjust the weights, $\alpha$ and $\beta$, for each teacher, guiding the student to selectively emphasize knowledge from the teacher with higher semantic value in each scenario. This dynamic weighting mechanism enables the student to effectively distill the most meaningful and relevant semantic knowledge, ultimately enhancing its performance and robustness across tasks.

\begin{itemize}
    \item \textbf{Low confidence in both teachers}: If both confidence scores fall significantly below a threshold $\delta$, both teachers are disregarded
    \item \textbf{Moderate confidence in both teachers}: Both teachers are assigned reduced weights, ensuring some influence without complete reliance.
    \item \textbf{Low confidence in one teacher}: Lower weight is assigned to the less confident teacher, prioritizing the more reliable one.
    \item \textbf{High confidence in both teachers}: Both teachers receive equal weighting.
\end{itemize}

\subsection{Machine Learning Models}

A recent study \cite{le2024board} provides a comprehensive analysis of ViTs performance and robustness in EO-IC, identifying EfficientViT and MobileViT as the two most effective models. Therefore, we select EfficientViT and MobileViT as our teacher models for this study. Technically, the EfficientViT model combines convolutional layers with local window attention mechanisms to optimize the balance between performance and computational efficiency, featuring approximately 4 million parameters \cite{liu2023efficientvit}. The MobileViT model integrates convolutional and transformer-based processing, using depthwise separable convolutions and self-attention mechanisms for high accuracy and efficient image classification  \cite{mehta2022separable}.

\begin{figure}[!t]
    \centering
    \includegraphics[scale=0.25]{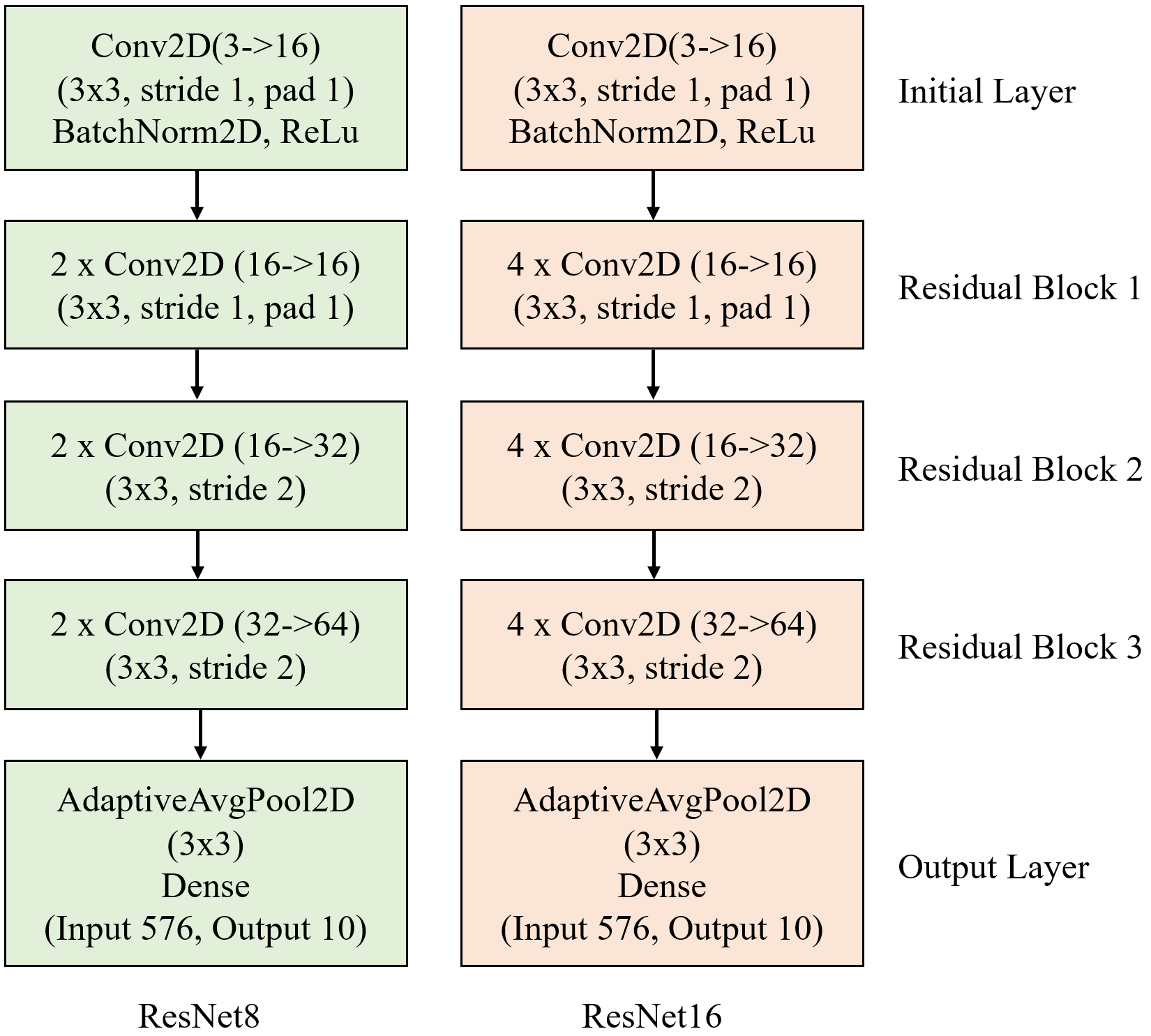}
    \captionsetup{font=small}
    \caption{Variants of ResNet-based student models network structure.}
    \label{fig:structure}
    \vspace{-6mm}
\end{figure}

ResNet has been shown to outperform standard CNNs for IC because of its ability to address the vanishing gradient problem through skip connections, allowing for deeper architectures and improved feature learning, as demonstrated in a comparative analysis \cite{mascarenhas2021comparison}. Additionally, ResNet is widely recognized for its effectiveness in KD training \cite{huang2021revisiting, guo2020online}. Therefore, we select a ResNet-based architecture as the student model for this study. We utilize two variants of ResNet—ResNet8 and ResNet16—as student models for KD, as shown in Fig. \ref{fig:structure}. These models are designed with progressively deeper architectures to balance performance with computational efficiency, making them suitable for deployment in resource-constrained environments, such as onboard satellite processing.

\begin{itemize}
    \item \textbf{ResNet8}: This lightweight ResNet variant consists of an initial convolutional layer followed by three residual blocks. Each block is structured to gradually increase the number of feature channels, from 16 to 64, through convolutional layers with either stride 1 or 2. The model concludes with an adaptive average pooling and dense layers. ResNet8’s shallow architecture makes it highly efficient for scenarios with limited computation power.

    \item \textbf{ResNet16}: This model builds on ResNet8’s structure by incorporating additional convolutional layers within each residual block, doubling the network's depth. The increased depth allows ResNet16 to capture more complex features, making it a more capable model for handling detailed image classification tasks. Like ResNet8, it uses an adaptive average pooling layer and a dense layer.
\end{itemize}

Both models are configured with Batch Normalization and ReLU activation functions to enhance training stability and convergence. Their structural differences offer a range of performance and efficiency trade-offs, providing flexibility for different use cases in the knowledge distillation framework.

\subsection{Evaluation Metrics}

To comprehensively evaluate the performance of our multiclass classification model across 10 classes, we employ three key metrics: accuracy, precision, and recall (sensitivity) \cite{goutte2005probabilistic}. These metrics are calculated for each class individually and then aggregated using macro-averaging to assess the model's performance as follows,
\vspace{-2mm}
\begin{align}
&\text{Accuracy} = \sum_{k=1}^K\frac{\text{TP}_k}{N}  \\ 
&\text{Precision} = \frac{1}{N} \sum_{i=1}^{K} N_k \frac{\text{TP}_k}{\text{TP}_k + \text{FP}_k}  \\ 
&\text{Recall} = \frac{1}{N} \sum_{i=1}^{K} N_k \frac{\text{TP}_k}{\text{TP}_k + \text{FN}_k} 
\end{align}
where $N$ is the total number of data points across all classes. $K$ is the total number of classes. $N_k$ is the number of data points in class $k$. $TP_k$ is True Positives, $FP_k$ is False Positives, $FN_k$ is False Negatives for class $k$, respectively. We use weighted precision and recall to ensure that each class is given equal importance, thereby providing a balanced evaluation of the model’s classification capabilities across the entire dataset. These macro-averaged evaluation metrics will select the best models in the final analysis.

\section{Experimental Setup}
All the experiments are conducted on GPU NVIDIA RTX™ 6000 Ada Generation, 48 GB GDDR6. Experiments were implemented using the Scikit-learn library \cite{scikit-learn}, and Pytorch. The data was divided into 70\% training and 30\% testing. In addition, we also applied batch normalization \cite{bjorck2018understanding} are employed for models' stability. Experiment parameters setting are summarized in Table \ref{table:experiment_parameters}.

\begin{table}[!t]
\centering
\caption{Experiment Parameters Setting}
\begin{tabular}{|l|c|}
\hline
\textbf{Parameter} & \textbf{Value} \\
\hline
Epoch & 50 \\
Batch size & 64 \\
Optimizer & AdamW \\
Learning rate & 0.00025 \\
Weight decay & 0.0005 \\
Scheduler & ReduceLRonPlateau \\
Threshold ($\delta$) & 0.6 \\
Temperature ($\tau$) & 5 \\
Min weight ($w_{min}$) & 0.1 \\
\hline
\end{tabular}
\label{table:experiment_parameters}
\end{table}

\section{Results and Discussions}

The performance comparison, as shown in Table \ref{tab:ResNet8_compare}, clearly illustrates the substantial improvements gained by applying KD, especially with a DualKD approach. Initially, employing KD with single teachers like EfficientViT and MobileViT boosts ResNet8’s accuracy from 87.76\% (base model) to 91.77\% and 91.06\%, respectively, demonstrating increases of approximately 4\% over the baseline. Precision and recall also reflect similar enhancements. However, the dual-teacher setup achieves the best results, increasing accuracy to 92.88\% - a total improvement of 5.12\% over the base model. Precision and recall also reach peak values of 93.07\% and 92.88\%, showing the most significant gains. This highlights how leveraging semantic insights from both teachers allows the model to capture richer, more nuanced features, resulting in a marked increase in performance across all metrics. 

\begin{table}[!t]
\centering
\caption{Performance Comparison of ResNet8}
\begin{tabular}{|l|c|c|c|}
\hline
\textbf{Model} & \textbf{Accuracy} $(\uparrow)$ & \textbf{Precision} $(\uparrow)$ & \textbf{Recall} $(\uparrow)$ \\ \hline
ResNet8 (Base) & 87.76 & 87.7 & 87.76 \\ \hline
ResNet8 (EfficientViT) & 91.77 & 91.74 & 91.77 \\ \hline
ResNet8 (MobileViT) & 91.06 & 91 & 91.06 \\ \hline
ResNet8 (Dual) & \textbf{92.88} & \textbf{93.07} & \textbf{92.88} \\ \hline
\end{tabular}
\label{tab:ResNet8_compare}
\captionsetup{font=footnotesize}
\caption*{\hspace{-5.5cm}\textbf{Bold} denotes the best values.}
\vspace{-2mm}
\end{table}

\begin{table}[!t]
\centering
\caption{Performance Comparison of ResNet16}
\begin{tabular}{|l|c|c|c|}
\hline
\textbf{Model} & \textbf{Accuracy} $(\uparrow)$ & \textbf{Precision} $(\uparrow)$ & \textbf{Recall} $(\uparrow)$ \\ \hline
ResNet16 (Base) & 92.9 & 92.91 & 92.9 \\ \hline
ResNet16 (EfficientViT) & 94.49 & 94.52 & 94.49 \\ \hline
ResNet16 (MobileViT) & 93.29 & 93.33 & 93.29 \\ \hline
ResNet16 (Dual) & \textbf{96.46} & \textbf{96.52} & \textbf{96.46} \\ \hline
\end{tabular}
\label{tab:ResNet16_compare}
\vspace{1mm} 
\captionsetup{font=footnotesize}
\caption*{\hspace{-5.5cm}\textbf{Bold} denotes the best values.}
\vspace{-5mm}
\end{table}

Similarly, the performance comparison for ResNet16 in Table \ref{tab:ResNet16_compare} demonstrates the significant gains achieved through KD, particularly with a DualKD approach. Utilizing single-teacher KD models like EfficientViT and MobileViT already enhances ResNet16’s performance, with EfficientViT increasing accuracy from 92.9\% (base model) to 94.49\%—an improvement of 1.59\%—and MobileViT achieving a slight increase to 93.29\%. However, the dual-teacher configuration yields the highest performance across all metrics, boosting accuracy to 96.46\%, a total improvement of 3.56\% over the base model. Precision and recall are similarly elevated, reaching 96.52\% and 96.46\%, respectively, marking the most substantial gains. Once again, this result underscores the DualKD strategy's effectiveness in enhancing ResNet16’s predictive capability, demonstrating that combining semantic insights from two teachers allows for a more comprehensive knowledge transfer, significantly improving the model’s overall performance.

However, despite these advancements, the student models still fall short of matching the performance levels of the teacher models, as summarized from \cite{le2024board}. The EfficientViT teacher model achieves an impressive accuracy of 98.76\%, precision of 98.77\%, and recall of 98.76\%. MobileViT, the highest-performing model in this comparison, reaches an accuracy, precision, and recall of 99.09\%. This difference highlights the gap between the student models and their teacher counterparts, illustrating that while KD with dual teachers substantially narrows the performance gap, the student models have yet to achieve the full predictive capacity exhibited by the teachers.

\begin{table*}[h!]
\centering
\caption{Model Comparison on Parameters, FLOPs, Size, Inference Time, and Power Consumption}
\begin{tabular}{|l|c|c|c|c|c|}
\hline
\textbf{Models} & \textbf{Total Parameters} ($\downarrow$) & \textbf{FLOPs} ($\downarrow$) & \textbf{Size (MB)} ($\downarrow$) & \textbf{Inference time (s)} ($\downarrow$) & \textbf{Power (W)} ($\downarrow$) \\ \hline
ResNet8 & \textbf{98,522} & \textbf{60,113,536} & \textbf{5.95} & \textbf{5.84} & \textbf{10.94 $\pm$ 0.83} \\ \hline
ResNet14 & 195,738 & 117,883,520 & 10.01 & 6.7 & 24.63 $\pm$ 1.63 \\ \hline
EfficientViT \cite{le2024board} & 3,964,804 & 203,533,056 & 38.19 & 10 & 29.04 $\pm$ 0.96 \\ \hline
MobileViT \cite{le2024board} & 4,393,971 & 1,843,303,424 & 259.30 & 16 & 79.23 $\pm$ 1.45 \\ \hline
\end{tabular}
\label{tab:complexity_compa}
\vspace{1.5mm} 
\captionsetup{font=footnotesize}
\caption*{\hspace{-11cm}\textbf{Bold} denotes the best values.}
\vspace{-6mm}
\end{table*}

Besides, ResNet8 achieves over 90\% in all evaluation metrics (accuracy, precision, and recall), meeting the required confidence level for reliable predictions. This strong performance underscores its effectiveness as a student model under the DualKD approach. The key advantage of using ResNet8 lies in its significantly lower complexity compared to its teacher models, EfficientViT and MobileViT, making it exceptionally suitable for deployment in resource-constrained environments.

As summarized in Table \ref{tab:complexity_compa}, with a parameter count of only 98,522, ResNet8 is 97.5\% smaller than EfficientViT (3,964,804 parameters) and 97.8\% smaller than MobileViT (4,393,971 parameters). It requires just 60,113,536 FLOPs, representing a 70.5\% reduction compared to EfficientViT and an impressive 96.7\% reduction compared to MobileViT. Additionally, ResNet8’s model size is only 5.95 MB, making it 84.4\% smaller than EfficientViT (38.19 MB) and 97.7\% smaller than MobileViT (259.30 MB). The inference time is equally optimized, with ResNet8 achieving 5.84 seconds, which is 41.6\% faster than EfficientViT and 63.5\% faster than MobileViT. Its power consumption is also considerably lower at 10.94 W $\pm$ 0.83 W, which is 62.3\% less than EfficientViT and 86.2\% less than MobileViT.  These reductions in complexity, size, and energy demands highlight ResNet8’s suitability for real-world applications where computational resources and power efficiency are critical. By maintaining high performance with low complexity, ResNet8 demonstrates the effectiveness of DualKD in creating a lightweight model that meets both predictive confidence and deployment constraints.

\begin{figure}[!t]
    \centering
    \includegraphics[width=\linewidth]{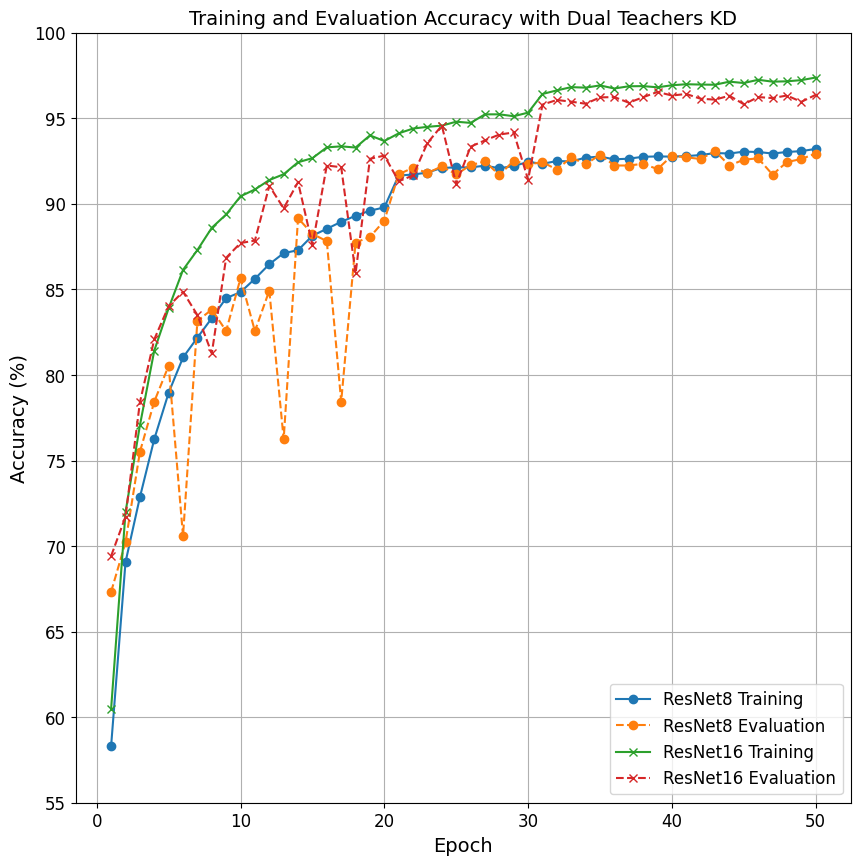}
    \captionsetup{font=small}
    \caption{Training and evaluation accuracy of ResNet variants with DualKD.}
    \label{fig:acc_performance}
    \vspace{-4mm}
\end{figure}

\begin{figure}[!t]
    \centering
    \includegraphics[width=\linewidth]{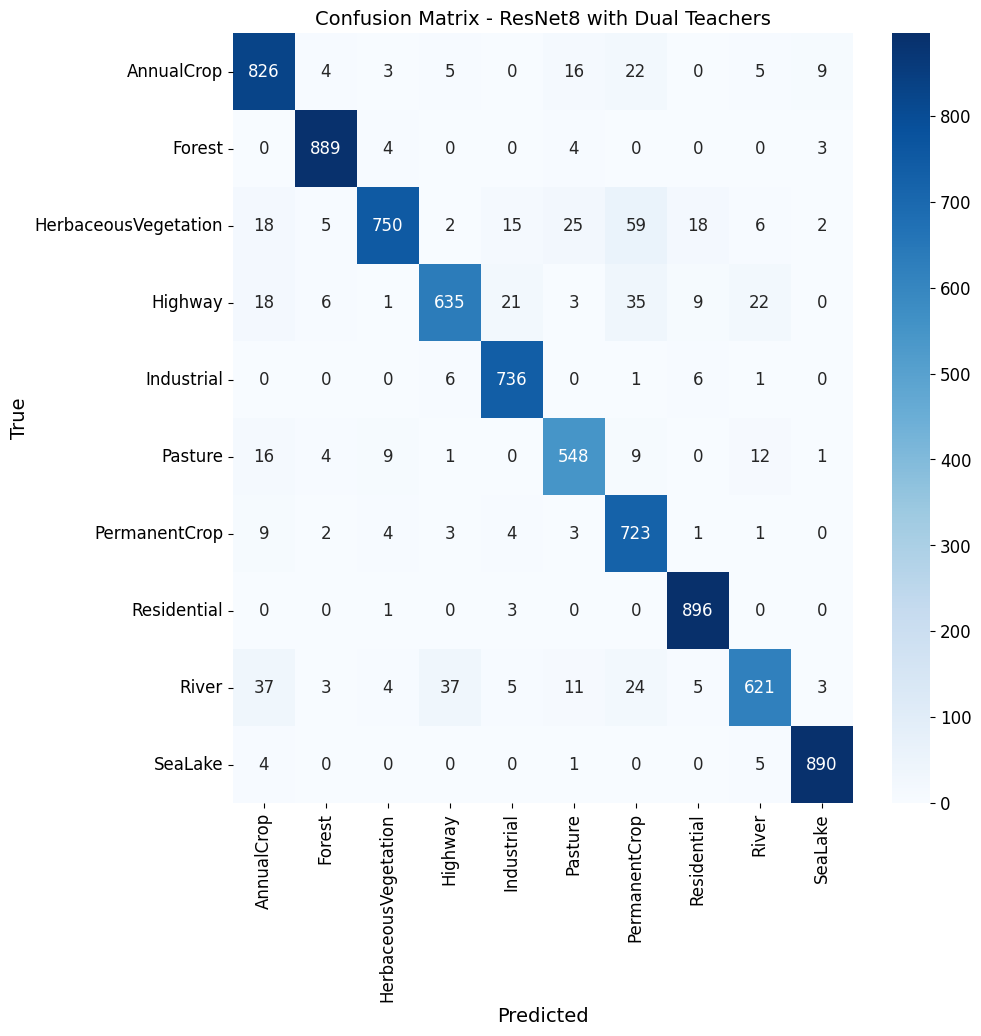}
    \includegraphics[width=\linewidth]{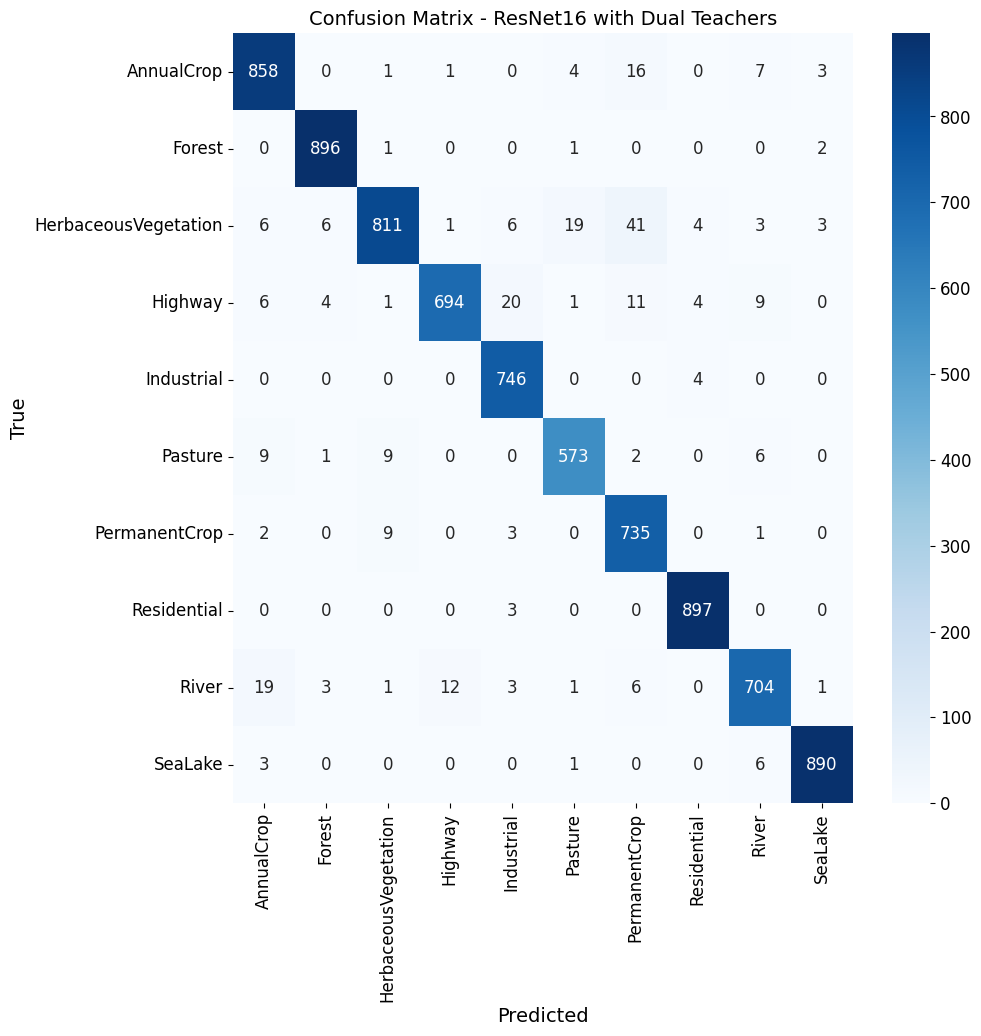}
    \vspace{-3mm}
    \captionsetup{font=small}
    \caption{Confusion matrix from ResNet8 (top) and ResNet16 (bottom) with DualKD.}
    \label{fig:conf_matrix}
\end{figure}

Figure \ref{fig:acc_performance} shows the training and evaluation accuracy for ResNet8 and ResNet16 with DualKD shows an initial phase of fluctuations, particularly in the evaluation metrics. During the first 20 epochs, both models experience notable variability, reflecting the model's adaptation to the dual-teacher signals. However, after 20 epochs, both ResNet8 and ResNet16 curves smooth out and converge, indicating stabilized learning and consistent improvement. Notably, ResNet8 exhibits a better convergence pattern than ResNet16, as evidenced by the smaller gap between its training and evaluation accuracy. This narrower difference suggests that ResNet8 generalizes more effectively, maintaining closer alignment between its training and evaluation performance. While ResNet16 ultimately achieves higher overall accuracy, comparing the confusion matrices of Fig. \ref{fig:conf_matrix}, it does so with a larger training-evaluation gap and at the cost of increased complexity—approximately double that of ResNet8. Given the marginal performance gain relative to its added computational cost, ResNet16 may not be worth the additional complexity, making ResNet8 the more efficient for applications requiring a balanced trade-off between accuracy and power consumption.

\section{Conclusions}

In conclusion, this study demonstrates the effectiveness of DualKD in enhancing the performance of lightweight student models, specifically ResNet8 and ResNet16. Both models achieve over 90\% accuracy, precision, and recall, meeting the required confidence level for reliable predictions and showing substantial improvements over baseline performances. ResNet8, in particular, strikes an optimal balance between high accuracy and efficiency, with significantly lower parameter counts, FLOPs, inference time, and power consumption. 

\section*{Acknowledgment}
This work was funded by the Luxembourg National Research Fund (FNR), with granted SENTRY project corresponding to grant reference C23/IS/18073708/SENTRY.

\bibliographystyle{IEEEtran}
\bibliography{IEEEabrv,Bibliography}
\end{document}